\documentclass[letterpaper]{article} 
\usepackage{aaai2026}  
\usepackage{times}  
\usepackage{helvet}  
\usepackage{courier}  
\usepackage[hyphens]{url}  
\usepackage{graphicx} 
\usepackage{natbib}  
\usepackage{caption} 
\usepackage{algorithm}
\usepackage{algorithmic}

\usepackage{newfloat}
\usepackage{listings}

\usepackage{amsmath}
\usepackage{multirow}
\usepackage{booktabs}
\usepackage{graphicx}
\usepackage{colortbl}

\DeclareCaptionStyle{ruled}{labelfont=normalfont,labelsep=colon,strut=off} 
\floatstyle{ruled}
\newfloat{listing}{tb}{lst}{}
\floatname{listing}{Listing}
\title{DreamSwapV: Mask-guided Subject Swapping for Any Customized Video Editing}

\author{
    Weitao Wang\textsuperscript{\rm 1,\rm 2}, Zichen Wang\textsuperscript{\rm 2}, Hongdeng Shen\textsuperscript{\rm 2}, Yulei Lu\textsuperscript{\rm 2}, Xirui Fan\textsuperscript{\rm 2}, Suhui Wu\textsuperscript{\rm 2},\\ Jun Zhang\textsuperscript{\rm 2}\thanks{Corresponding Authors.}, Haoqian Wang\textsuperscript{\rm 1}\footnotemark[1], Hao Zhang\textsuperscript{\rm 2}\thanks{Project Leader.}
}
\affiliations{
    \textsuperscript{\rm 1}Tsinghua University~~
    \textsuperscript{\rm 2}ByteDance\\
}

\nocopyright

\usepackage{bibentry}

\begin{document}

\maketitle

\begin{abstract}
With the rapid progress of video generation, demand for customized video editing is surging, where subject swapping constitutes a key component yet remains under-explored. Prevailing swapping approaches either specialize in narrow domains—such as human-body animation or hand-object interaction—or rely on some indirect editing paradigm or ambiguous text prompts that compromise final fidelity. In this paper, we propose DreamSwapV, a mask-guided, subject-agnostic, end-to-end framework that swaps any subject in any video for customization with a user-specified mask and reference image. To inject fine-grained guidance, we introduce multiple conditions and a dedicated condition fusion module that integrates them efficiently. In addition, an adaptive mask strategy is designed to accommodate subjects of varying scales and attributes, further improving interactions between the swapped subject and its surrounding context. Through our elaborate two-phase dataset construction and training scheme, our DreamSwapV outperforms existing methods, as validated by comprehensive experiments on VBench indicators and our first introduced DreamSwapV-Benchmark.
\end{abstract}
\vspace{-1.1em}

\section{Introduction}

The recent development of video generation technologies is enabling a broader audience to engage in customized content creation, also stimulating demand for editing real or generated videos. Users increasingly desire to introduce specific subjects into videos following the original motion trajectory, necessitating the video subject swapping task. Given a source subject in the video and a target subject to be inserted, the ideal outcome is that the target is seamlessly integrated into the original video, preserving its own appearance details, following the source motion, and interacting naturally with the surrounding context. 

\begin{figure}[t]
\centering
\includegraphics[width=0.45\textwidth]{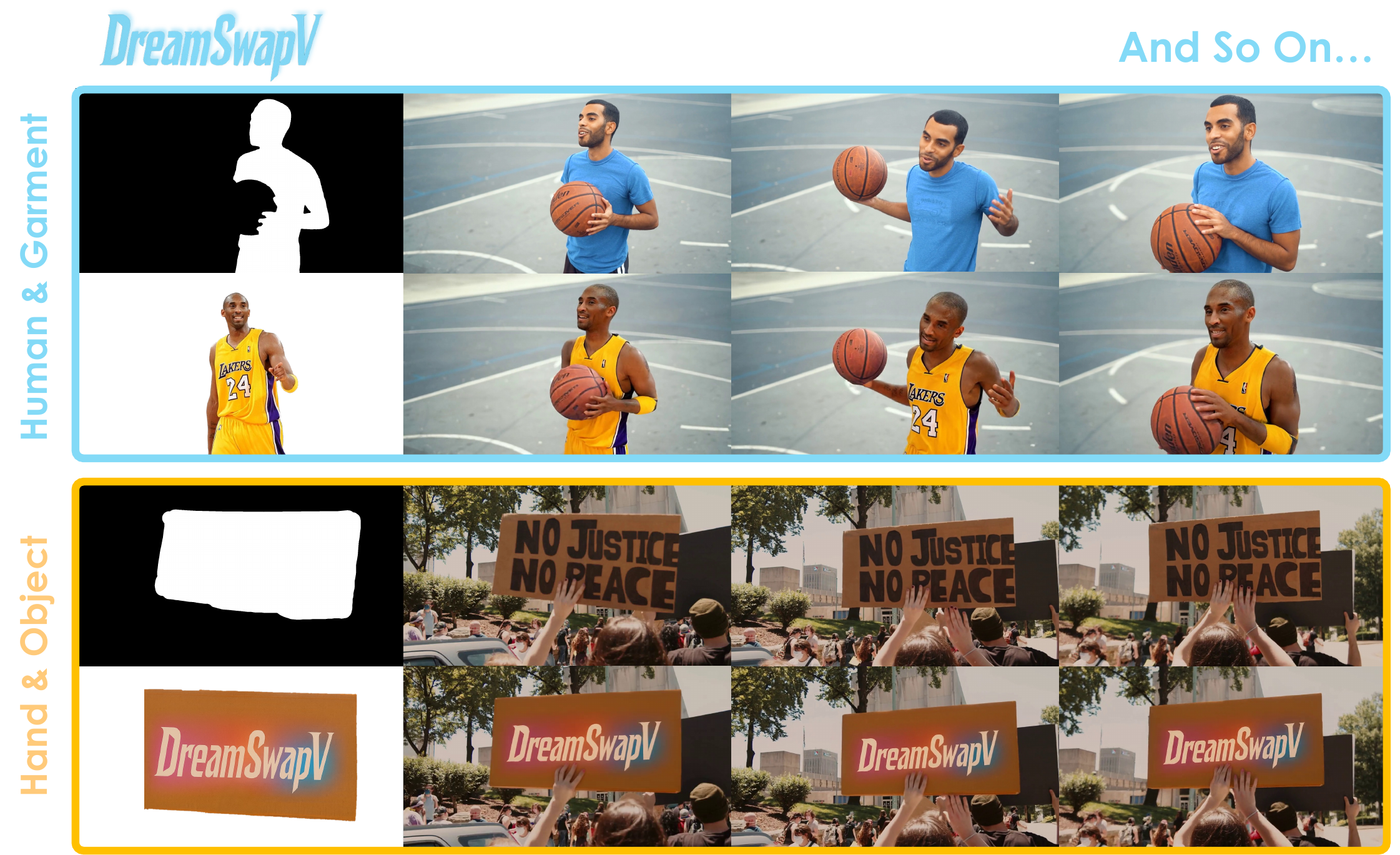} 
\caption{Our DreamSwapV can swap any subject in any video with a user-specified mask and reference image.}
\label{first}
\vspace{-1.3em}
\end{figure}

However, existing video editing methods exhibit critical limitations: On one hand, domain-specific approaches are confined to particular subjects: animation techniques like MagicAnimate \cite{magicanimate} and Animate Anyone 2 \cite{animate} only support swapping human characters; Human-Object Interaction (HOI) methods like AnchorCrafter \cite{anchorcrafter} and DreamActor-H1 \cite{dreamactor} are limited to specific scenarios (e.g. livestreaming, E-commerce) for swapping handheld objects; thereby restricting their broader applicability. On the other hand, general-purpose video editing methods also face several challenges: (1) Tuning-free techniques \cite{anyv2v, videograin} manipulate attention features to swap subjects, but struggle to restore fine details and largely ignore interactions with subjects; (2) Tuning-based methods inject the target subject either via text prompts like VideoPainter \cite{videopainter} or learned LoRA from several subject concepts like VideoSwap \cite{videoswap}; the former lacks detail fidelity, whereas the latter is computationally demanding and indirect in reference injection; (3) Recently, unified video customization frameworks \cite{vace, hunyuancustom} are emerging, aiming for versatility across multiple customization tasks, including video-driven subject swapping. Yet their all-in-one design sacrifices identity consistency and interaction realism in the subject swapping scenario.

To address these limitations, we present DreamSwapV, a mask-guided, subject-agnostic framework for video subject swapping. Given a source video, a user-specified mask localizing the source subject, and a reference image of the target subject, we can perform end-to-end swapping of any subject in any video. Unlike conventional editing paradigms that focus on \emph{generating} or \emph{editing} an external object into the scene, we recast swapping as an \emph{inpainting} task: we encourage the model to \emph{recover} the appearance of the target subject as if it inherently belonged within the missing masked region of the source video, yielding a direct and intuitive training and inference procedure.

Specifically, we first construct a task-oriented dataset based on HumanVID \cite{humanvid} and introduce auxiliary conditions for better video control. A dedicated condition fusion module then effectively integrates these conditions for model's better understanding of both video content and task objectives. Furthermore, an adaptive mask strategy accommodates subjects of varying scales and attributes, mitigating shape leakage while enhancing subject–context interaction. Lastly, we establish DreamSwapV-Benchmark—the first benchmark tailored to video subject swapping—and evaluate our method along with 5 indicators inherited from VBench \cite{vbench}, confirming the robustness and superior performance of DreamSwapV.

Our contributions are summarized as follows:

\begin{itemize}
    \item We propose DreamSwapV, the first end-to-end framework dedicated to the video subject swapping, specifically focusing on generic subject-context interaction.
    \item Our novel condition fusion module and adaptive mask strategy enable handling diverse subjects, enhancing contextual understanding, and improving video realism.
    \item Through a two-stage dataset construction and training scheme, we outperform existing methods on VBench indicators and first introduced DreamSwapV-Benchmark.
\end{itemize}

\section{Related Work}

\subsection{Video Generation and Editing}

Video generation is a pivotal component within the Artificial Intelligence Generated Content (AIGC) domain, realizing human creativity in the video medium. The field has undergone rapid evolution, from initial GAN-based methods \cite{vgan, mocogan}, through subsequent diffusion models \cite{vdm, makeavideo}, to the current state-of-the-art Diffusion Transformers (DiT) \cite{dit}. At present, leading commercial models \cite{Keling, Hailuo, vidu} and open-source models \cite{cogvideox, hunyuanvideo} can leverage billion-scale architectures to generate high-fidelity videos at 720p-1080p resolutions.

Concurrently, editing techniques for real or generated videos are also advancing vigorously. Tune-A-Video \cite{tune} pioneers video editing with latent diffusion models via one-shot tuning, followed by \cite{fatezero, flatten, controlvideo}, enhancing capabilities in appearance detail, temporal consistency and motion control. With the advent of video foundation models, an increasing number of editing methods are based on video diffusion models \cite{videoswap, videograin} or DiT \cite{videopainter, videoanydoor}, providing more precise and diverse video editing. Our DreamSwapV builds upon the cutting-edge Wan2.1 DiT model \cite{wan} and adapts it specifically to the video subject swapping task, with our condition fusion module injecting additional conditioning information and a carefully designed training scheme.

\subsection{Video Customization}

The freedom and creativity of personalized content creation facilitates the development of video customization, which can be broadly classified into two paradigms. \textbf{Subject concepts-driven methods} learn subject-specific adapters \cite{adapter} or LoRA modules \cite{lora} from multiple reference images, embedding the subject concepts into the text space. DreamBooth \cite{dreambooth} and Textual Inversion \cite{textual} pioneer this approach in the image domain and later extend to video customization by methods like \cite{still, customcrafter, videobooth}. While effective for identity preservation, the requirement of per-subject finetuning hampers real-time use, motivating the emergence of end-to-end methods. 

\textbf{End-to-end customization} typically employs an additional conditioning network to inject user-specified information, similar to the idea of ControlNet \cite{controlnet}. Animation techniques \cite{animate} design a ReferenceNet specifically for injecting detailed character appearance. Unified video customization frameworks \cite{vace, hunyuancustom} broaden the scope to encompass comprehensive customization capabilities, including text-to-video (T2V), reference-to-video (R2V), video-to-video (V2V) and masked-video-to-video (MV2V).

\subsection{Video Subject Swapping and Inpainting}

Video subject swapping is a derived sub-task of video customization, focusing on subject-driven editing and MV2V, which has garnered significant attention due to rising user demand. Some general-purpose video editing methods \cite{anyv2v, videograin, videoswap} inherently support video subject swapping, which operate in a \emph{generating} or \emph{editing} paradigm , rather than \emph{video inpainting}. 

Video inpainting \cite{deepvp, propainter} itself is a long-established research field, originally concerned with restoring missing or damaged content in videos. Crucially, when the masked region represents an unknown external subject, video inpainting techniques can effectively \emph{recover} this novel object, thereby achieving the functional goal of video subject swapping—as demonstrated by VideoPainter \cite{videopainter} and the MV2V capabilities of unified customization frameworks \cite{vace, hunyuancustom}. Our DreamSwapV directly inherits and builds upon this core insight, by training a dedicated, end-to-end model specifically designed to achieve generic and high-quality video subject swapping in a video inpainting manner.
\begin{figure*}[t]
\centering
\includegraphics[width=0.95\textwidth]{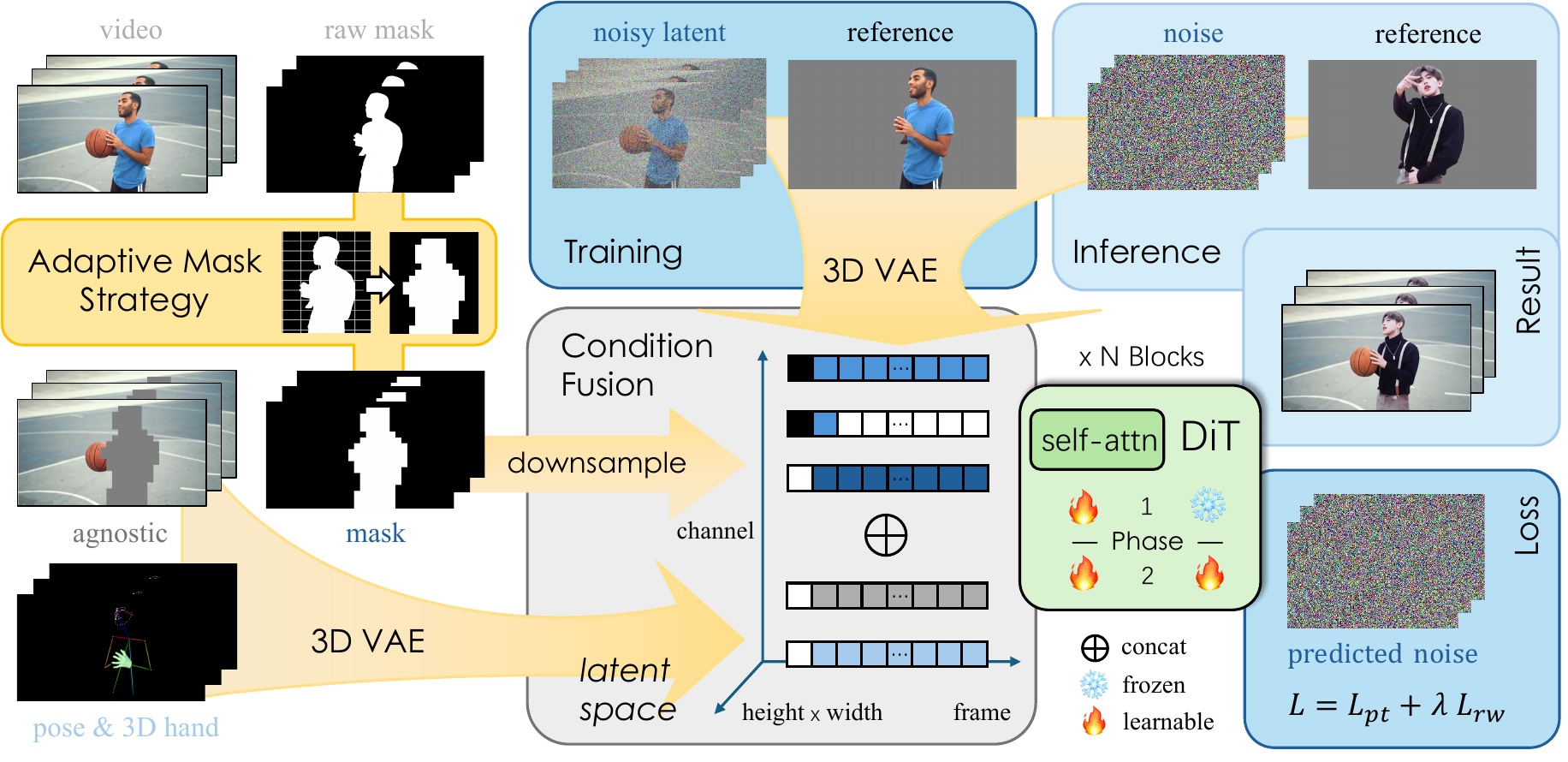} 
\caption{An overview of our whole framework. We first construct our pre-training dataset with video, raw mask, pose and 3D hand sequences (Sec. 3.1). Then our condition fusion module integrates these conditions together efficiently, ensuring strict spatial-temporal alignment across all conditioning signals (Sec. 3.2). To better handle subject-context interactions and achieve generic subject swapping, we propose an adaptive mask strategy processing raw mask sequences (Sec. 3.3). After a two-phase training scheme (Sec. 3.4), our DreamSwapV delivers state-of-the-art subject swapping results at inference time.}
\label{pipe}
\vspace{-1em}
\end{figure*}
\section{Method}

\subsection{Task Overview}

As fully discussed above, we cast video subject swapping as a special application of video inpainting: a user-specified first-frame mask $\mathbf{m}_0^s$ delimits a region to be swapped, and a corresponding reference image $\mathbf{r}^s$ supplies the appearance of the new subject. Under this setting, the task can also transfer to (i) \emph{standard video inpainting} when the reference is absent, or (ii) \emph{video addition} when the masked region does not contain a pre-existing subject. In this work, we mainly focus on the canonical video subject swapping task—namely, the mask encloses a single subject and the reference depicts another subject of similar category.

\noindent\textbf{Dataset Construction.} To construct a dataset tailored for our task, we leverage HumanVID \cite{humanvid} as the foundational resource. Beyond its human-centric focus, HumanVID offers extensive coverage of human-object interactions and diverse subject categories, aligning with our requirement for generic subject swapping. 

Then we establish a rigorous data processing pipeline (video subject caption $\Rightarrow$ per-frame subject mask tracking $\Rightarrow$ quality filtering $\Rightarrow$ distribution standardization), as detailed in Sec. 3.4. At last, each dataset sample comprises (i) the raw video $\mathbf{V}=\{\mathbf{v}_t\}^T_{t=1}$, (ii) pose \& 3D hand sequences $\mathbf{P}=\{\mathbf{p}_t\}^T_{t=1}$, and (iii) per-subject masks $\mathbf{M^s}=\{\mathbf{m}_t^s\}^T_{t=1}$, where $s$ means the specific subject.

\noindent\textbf{Training and Inference.} During training, the reference image is collected by $\mathbf{r'}=\mathbf{v}_i\odot\mathbf{m}_i$, where $i$ is a randomly selected frame and $\odot$ denotes element-wise product, i.e., extracting the subject marked by the mask from the video frame as the training reference image. The model is thus trained to \emph{recover} the original video by inpainting the masked region using this extracted reference image, encouraging faithful appearance recovery and seamless interaction with the scene, which can be expressed as follows:
\begin{equation}
\text{Loss}=L_{pt}(\mathbf{V}, \mathbf{V'}), \mathbf{V'}=f_\theta(\mathbf{M^s}, \mathbf{P},\mathbf{r'})
\end{equation}

where $L_{pt}$ represents the pre-trained loss, whose final calculation will be supplemented in Sec. 3.4, and $f_\theta$ depicts the \emph{recovering} procedure of our model with weight $\theta$.

At inference time, the model treats an external reference image $\mathbf{r^s}$ as if it were the subject originally extracted from the mask region, leveraging \emph{recovering} capability learned during training. Thus we can obtain the customized video $\mathbf{V'}$ after our model's subject swapping: $\mathbf{V'}=f_\theta(\mathbf{M^s}, \mathbf{P},\mathbf{r^s})$.

\subsection{Condition Fusion Module}

The subject swapping task inherently requires the model to distinguish between regions requiring to be swapped and those to be preserved, which is relatively straightforward to learn with the binary mask serves as a key condition. Building on this, the more challenging task is (i) accurately recovering the motion trajectory of the swapped subject, and (ii) handling the mask boundary interactions between the swapped subject and its surrounding context.

For dynamic subjects (e.g., humans and animals), self-occlusion frequently occurs during movement, necessitating the pose estimation as an additional condition to provide the subject's motion information. For static subjects (e.g., objects), the motion pattern from camera movement or external forces is simpler, and usually can be inferred from mask shape changes. However, the realism of interactions between these objects and external forces (e.g. hand manipulations) critically impacts visual quality. To better handle these interactions, we introduce 3D hand estimation as a complementary signal, providing hand-object interaction information. The pose and 3D hand estimation are combined as a unified temporal sequence $\mathbf{P}$ to offer their auxiliary control.

Along with the necessary mask sequence $\mathbf{M^s}$, the masked video sequence $\mathbf{A^s=\mathbf{V}\odot(1-\mathbf{M^s})}$ (often referred to as agnostic) and the reference image $\mathbf{r}^s$, we thus incorporate four additional conditions along with the noisy video input. Efficiently coupling this multi-modal information is essential to overall performance, where we address this challenge through a dedicated condition fusion module.

\noindent\textbf{Latent Space Projection.} All input sequences except the binary masks are encoded into a shared latent space using a pre-trained 3D VAE. The shape of these latents becomes [batch, frame, channel, height, width] = [b, f, c, h, w], with their temporal dimension (f) compressed by 4 and spatial dimensions (h, w) compressed by 8. The reference latent remains f = 1, and the remaining sequences are scaled to f = (frames - 1) // 4 + 1. Next, the first frame of the noisy video latent is extracted separately to form a dummy reference latent, with its subsequent frames zero-padded to match the temporal length of the noisy video latent. This latent explicitly represents the first frame of the whole video sequence, enabling the model to focus on the extension of the first frame, which provides an interface for extrapolating the video length and first frame reference as detailed in Sec. 3.4.

\noindent\textbf{Reference Information Injection.} Our reference latent is concatenated along the temporal dimension (f) to both the noisy video latent and the dummy reference latent. We critically compare the following injection alternatives:

(i) \emph{Channel concatenation:} It is a straightforward idea to concatenate the zero-padded or all-copied reference latent along the channel dimension, just as we do with the dummy reference latent and other latents. However, this may disrupts the spatial-temporal feature alignment across frames since the reference latent represents a global signal applicable to all frames, not just frame 0, leading to confused learning and impaired detail injection.

(ii) \emph{Cross-attention:} Extracting other features (like CLIP features) instead of 3D VAE latents from the reference for cross-attention calculation suffers from limited detail extraction from encoders like CLIP. Cross-attention is better suited for semantic guidance (e.g., text embeddings), not high-fidelity appearance injection.

(iii) \emph{ReferenceNet:} In animation methods, an additional ReferenceNet is typically used to implicitly inject reference features. However, as demonstrated by UniAnimate \cite{unianimate}, such operation introduces parameter redundancy and feature space misalignment between the reference image and noisy video latent, which can be replaced.

\begin{figure}[t]
\centering
\includegraphics[width=0.48\textwidth]{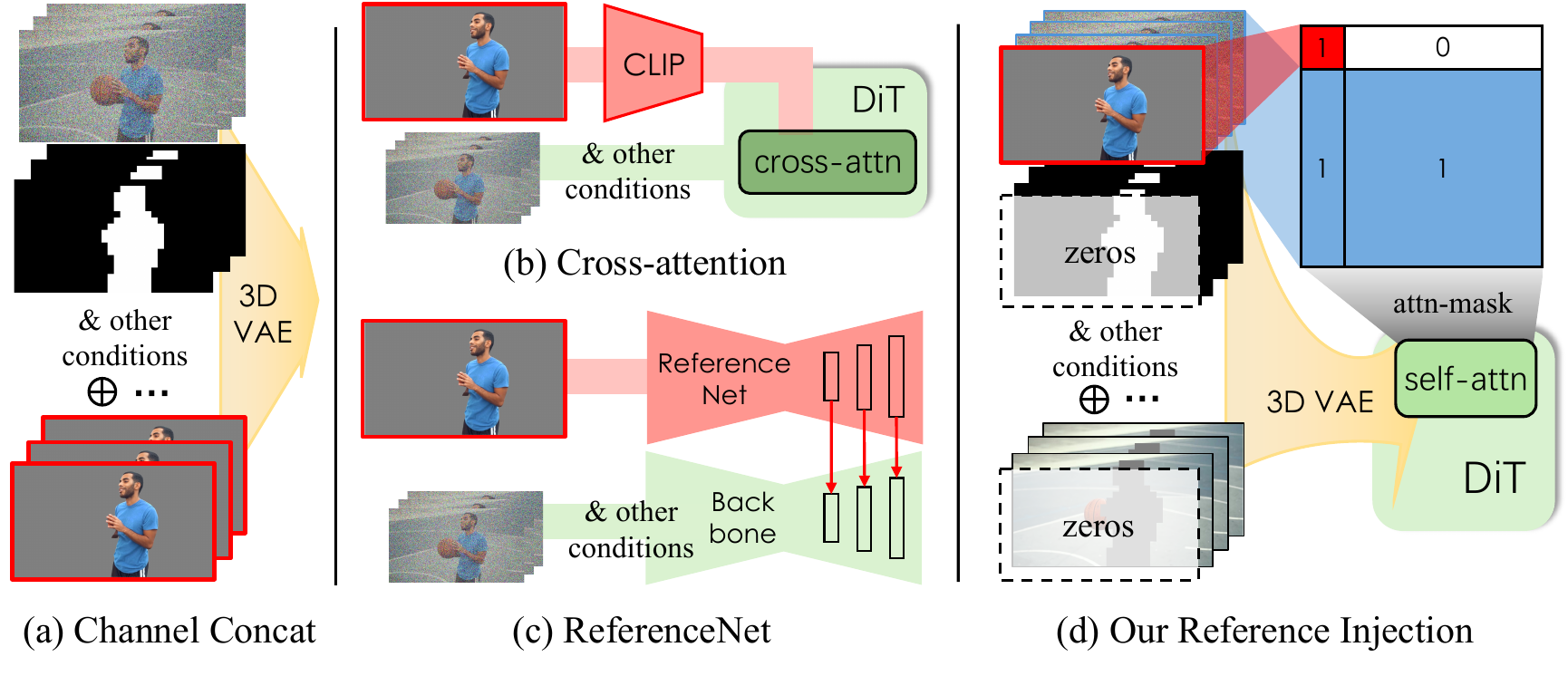} 
\vspace{-1.5em}
\caption{Schematic comparison between reference information injection alternatives (a)-(c) and our method (d).}
\label{ref}
\vspace{-1em}
\end{figure}

Therefore, we decide to use frame-level concatenation by extending the token length, enabling seamless reference injection. Since the reference is only used for feature extraction and not for prediction, in the self-attention mechanism, the video attends to the reference while the reference only attends to itself (using KV cache), as shown in the attention mask of Figure \ref{ref}. The reference latent is also excluded from the loss computation, thus achieve the same effect as ReferenceNet in a more concise temporal concatenation manner.

\noindent\textbf{Mask Conditioning and Final Fusion.} The binary mask sequence does not require 3D VAE encoding. To match the spatial-temporal dimensions of the other condition latents, the masks are processed by concatenating every four frames along the channel dimension, followed by an 8x downsampling in spatial dimensions, resulting in a [b, (f - 1) // 4 + 1, 4, h // 8, w // 8] shape. Finally, the mask, agnostic latent, and pose latent are zero-padded on the reference frame, and all the latents are concatenated along the channel dimension to fuse as the final $\mathbf{model\_input}$.

In summary, our condition fusion module ensures strict spatial-temporal alignment across all conditioning signals. For each frame in the temporal dimension, corresponding conditions of the frame are concatenated along the channel, with additional reference information injected globally via token extension, simplifying the model's learning.
\subsection{Adaptive Mask Strategy}

As a mask-guided task, the handling of the mask is crucial in the video subject swapping. \emph{Overly precise} masks may cause the model to rely on the shape of the masks during training, leading to poor generalizability (often referred to as \emph{shape leakage}) when swapping subjects with significant domain gaps (e.g. square box $\Rightarrow$ spherical ball). On the other hand, \emph{overly coarse} masks may confuse the model's learning, resulting in artifacts and less refined details. The proper handling of the mask boundary also decisively affects the subject-context interactions, and finally, governs the overall naturalness and visual quality of the subject swapping.

Prior work like Animate Anyone 2 \cite{animate} address such dilemma using a grid-based mask augmentation strategy: the bounding box of the mask is divided into $k_h\times k_w$ non-overlapping blocks; any block containing mask pixels is dilated, effectively blurring the precise mask boundary to reduce \emph{shape leakage}. While Animate Anyone 2 is designed specifically for human swapping (animation), our goal is to perform generic subject swapping across diverse scales and attributes. Directly applying their augmentation strategy would result in both small items (e.g. accessories, handheld objects) and relatively large objects (e.g. humans, furniture, background) sharing the same augmentation, which is suboptimal for both. Therefore, we need to explore a more generalizable mask augmentation strategy.

\noindent\textbf{Adaptive Grid Sizing.} To begin with, we randomly use a bounding box augmentation with 30\% probability, representing the largest shape augmentation (coarsest mask).  In the remaining 70\% of cases, we inherit the grid-augmentation strategy of Animate Anyone 2, dividing the whole frame rather than the bounding box and making the grid size inversely proportional to the subject scale:
\begin{equation}
\underbrace{
  \left\{
    \begin{aligned}
      k_h^{train}&=\text{rand}(1, h) \\ k_h^{inf}&=h//10 
    \end{aligned}
  \right.
  }_{\text{Animate Anyone 2}}
  \Rightarrow
\underbrace{
  \left\{
    \begin{aligned}
      K_h^{train}&=\text{bbox}\_h//\text{rand}(h_1, h_2)\\ K_h^{inf}&=\text{bbox}\_h //h_3
    \end{aligned}
  \right.
  }_{\text{Our Adaptive Grid Sizing}}
\label{eq}
\end{equation}

where $k_h^{train}, k_h^{inf}$ denote the number of vertical blocks into which Animate Anyone 2 divides the mask's bounding box during training and inference; $K_h^{train}, K_h^{inf}$ calculate our adaptive vertical block number, based on the bounding box height $\text{bbox}\_h$ and hyper-parameters $h_1, h_2, h_3$. 

The horizontal counts $k_w, K_w$ are computed analogously. Intuitively, the larger the subject, the larger $\text{bbox}\_h\times \text{bbox}\_w$, and hence the larger $K_h\times K_w$, which yields a finer grid, better controlling their movement (and the larger subjects' swapping rarely involves \emph{shape leakage}). Conversely, small subjects receive coarser masks which accommodate more diverse swapping (e.g., handheld objects to other categories). We empirically find that this adaptive strategy improves the generalization for small subjects and enhances the edge and motion control over larger subjects.

\noindent\textbf{Extra Shape Augmentation.} Animate Anyone 2 further suggests introducing random scaling for the mask region to avoid that the augmented mask is always larger than the original character. We do not view this as a problem—on the contrary, we prefer the mask to always exceed the subject to be swapped, simplifying the model's task objective. We argue that guiding the model to learn how to reasonably fill in the subject and its surroundings within a slightly larger mask is easier than relying on it to speculate whether the mask is larger than the swapped region. The latter often leads to hallucinations and confusions in the model's reasoning. 

Based on this, we propose to further address those hallucinations (e.g., characters growing incorrect black hair), which indicates the model's insufficient ability to reasonably fill in the non-subject surroundings. During training, we occasionally perform extra shape augmentation by introducing randomly generated shapes (e.g., circle, triangle, rectangle) onto the edges of the existing grid augmented mask. This further decouples the subject from the shape of its mask, making it more explicit to the model that not every masked pixel belongs to the subject. As a result, our model can better handle the mask boundary with appropriate background content, thereby improving subject-context interactions.

Overall, our adaptive mask strategy provides a generic mask augmentation scheme that adapts to subjects of various scales and attributes. Combined with the pose and 3D hand conditions, the grid-based and extra shape augmentations further supply the model with better subject-context understanding, producing smoother and more natural boundaries between the swapped subject and its surroundings.
\begin{figure}[t]
\centering
\includegraphics[width=0.46\textwidth]{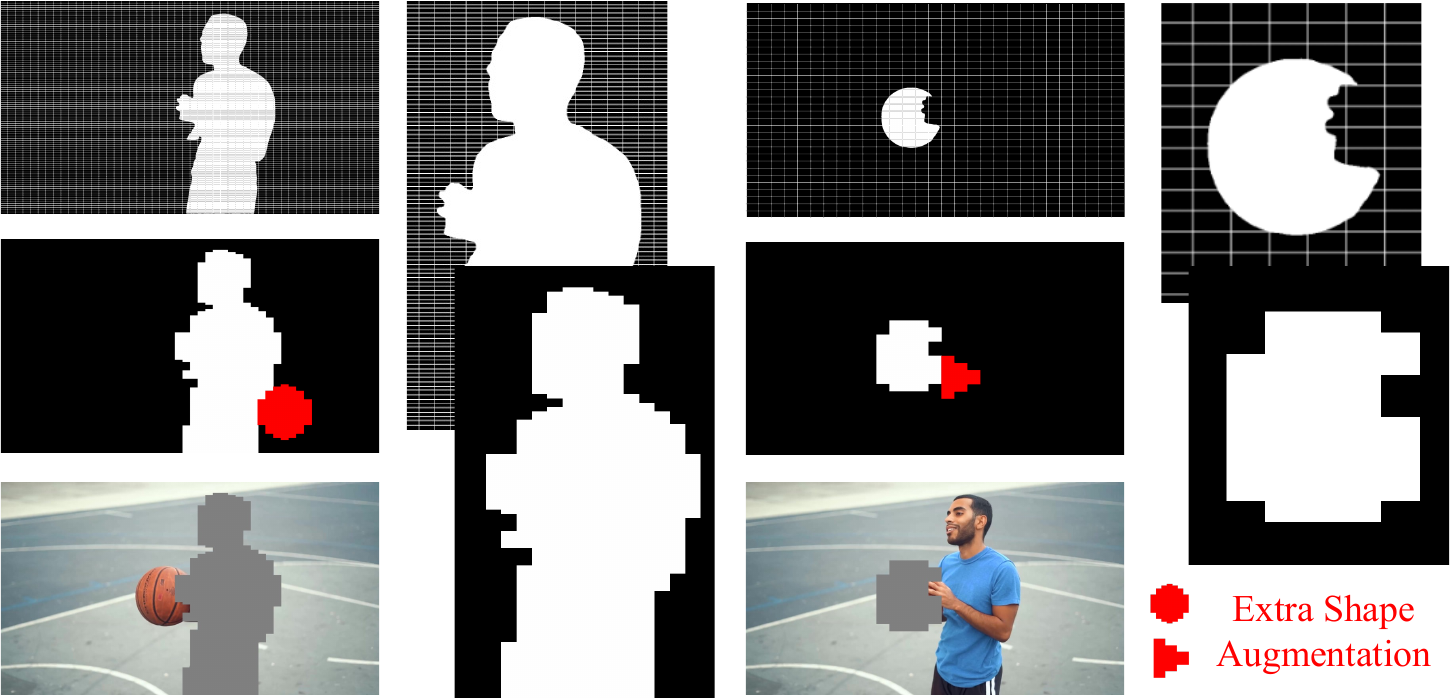} 
\caption{Illustration of our adaptive grid sizing and extra shape augmentation on subjects of different scales.}
\label{text}
\vspace{-1em}
\end{figure}
\subsection{Implementation Details}

We list several technical details that, while not totally novel, proved essential to the final performance of our model.

\begin{table*}[]
\centering
\small
\setlength{\tabcolsep}{1.4pt}
\renewcommand{\arraystretch}{1.2}
\begin{tabular}{l|
>{\centering\arraybackslash}m{1.2cm} >{\centering\arraybackslash}m{1.2cm} >{\centering\arraybackslash}m{1.2cm} >{\centering\arraybackslash}m{1.2cm} >{\centering\arraybackslash}m{1.2cm} >{\centering\arraybackslash}m{1.2cm} >{\centering\arraybackslash}m{1.2cm} >{\centering\arraybackslash}m{1.2cm} >{\centering\arraybackslash}m{1.2cm}|
>{\centering\arraybackslash}m{1.2cm} >{\centering\arraybackslash}m{1.2cm} >{\centering\arraybackslash}m{1.2cm}}
\toprule
\multirow{2}{*}{Method} &
\multicolumn{9}{c|}{\textbf{Video Quality \& Video Consistency}} &
\multicolumn{3}{c}{\textbf{User Study}} \\ ~~~~~~~~/~~ Metrics
& \rotatebox{90}{\shortstack[c]{\textcolor{pink!50!white}{\large\textbullet} Subject\\Consistency}}
& \rotatebox{90}{\shortstack[c]{\textcolor{yellow!80!white}{\large\textbullet} Background\\Consistency}}
& \rotatebox{90}{\shortstack[c]{\textcolor{orange!50!white}{\large\textbullet} Motion\\Smoothness}}
& \rotatebox{90}{\shortstack[c]{\textcolor{red!50!white}{\large\textbullet} Dynamic\\Degree}}
& \rotatebox{90}{\shortstack[c]{\textcolor{lime!70!white}{\large\textbullet} Aesthetic\\Quality}}
& \rotatebox{90}{\shortstack[c]{\textcolor{cyan!50!white}{\large\textbullet} VBench\\Average}}
& \rotatebox{90}{\shortstack[c]{\textcolor{blue!50!white}{\large\textbullet} Reference\\Appearance}}
& \rotatebox{90}{\shortstack[c]{\textcolor{violet!50!white}{\large\textbullet} Background\\Preservation}}
& \rotatebox{90}{\shortstack[c]{\textcolor{purple!40!white}{\large\textbullet} Total\\Average}}
& \rotatebox{90}{\shortstack[c]{\textcolor{teal!40!white}{\large\textbullet} Reference\\Detail}}
& \rotatebox{90}{\shortstack[c]{\textcolor{brown!30!white}{\large\textbullet} Subject\\Interaction}}
& \rotatebox{90}{\shortstack[c]{\textcolor{gray!30!white}{\large\textbullet} Visual\\Fidelity}} \\

\midrule
AnyV2V & 90.03\% & 91.35\% & 98.60\% & 47.90\% & 51.79\% & 75.93\% & 34.70\% & 42.71\%  & 65.30\% & 0.87 & 0.65 & 0.42 \\
VACE & \underline{96.15\%} & \underline{95.03\%} & 99.29\% & 27.54\% & 56.95\% & 74.99\% & 39.66\% & 47.46\% & 66.01\% & 2.09 & 2.31 & 2.46 \\
HunyuanCustom & 95.83\% & 94.96\% & 99.17\% & 43.11\% & \textbf{57.78\%} & 78.17\% & 41.33\% & \underline{48.14\%} & 68.61\% & 2.17 & 2.22 & 2.13 \\
Kling 1.6 & 95.36\% & \textbf{96.57\%} & \textbf{99.45\%} & \underline{50.33\%} & \underline{57.26\%} & \underline{79.79\%} & \underline{42.27\%} & 39.17\% & \underline{68.63\%} & \underline{3.04} & \underline{2.89} & \underline{3.14} \\
\rowcolor{gray!15}
\textbf{DreamSwapV} & \textbf{96.41\%} & 94.26\% & \underline{99.31\%} & \textbf{55.69\%} & 56.52\% & \textbf{80.44\%} & \textbf{45.22\%} & \textbf{52.49\%} & \textbf{71.41\%} & \textbf{3.35} & \textbf{3.39} & \textbf{3.32} \\
\bottomrule
\end{tabular}
\caption{Quantitative comparison of subject swapping methods on DreamSwapV-Benchmark with VBench indicators and user study. The \textbf{bold} and \underline{underline} are the top-1 and top-2 results under certain metrics, respectively, and \colorbox{gray!15}{gray} marks our method.}
\label{table}
\vspace{-1em}
\end{table*}

\noindent\textbf{Data Preprocessing and Filtering.} Based on HumanVID, TikTok-VFM-7B is employed to caption prominent subjects of each video, followed by TrackingSAM \cite{trackingsam} performing per-frame tracking and segmentation of each subject. The obtained subject mask sequences then undergo strict quality filtering based on three criteria below: \emph{(i) the mask area ratio} (relative to the full frame), \emph{(ii) the temporal coverage} (of total frames), and \emph{(iii) the motion amplitude} (displacement across frames). To prevent our model from being overly biased towards any specific subject category, we further standardize the distribution of subjects in the dataset by filtering out excessively repetitive instances, ensuring a balanced diversity of subject categories. 

Following these steps, we finally construct our pre-training dataset containing 8160 videos and 16219 subject instances (4701 \emph{humans} : 1045 \emph{garments} : 5477 \emph{small objects} : 4996 \emph{large objects} $\approx$ 1 : 0.2 : 1 : 1, where \emph{small objects} include highly human-related items like accessories and handheld objects, and \emph{large objects} include other items like furniture, buildings, and background). Pose and 3D hand sequences are detected together by DWPose \cite{dwpose} and Hamer \cite{hamer} per-frame.

\noindent\textbf{Reference Augmentation.} During pre-training, the reference image is always extracted from the source video, maintaining its perfect scale and luminance. This risks the model learning a trivial \emph{copy-paste} behavior (we call it as \emph{reference leakage}, also noted in HunyuanCustom), impairing its ability to handle external real-world references with significant domain gaps (e.g. scale, brightness, or pose mismatches). To mitigate this, we augment the training reference with random scaling, rotation, flipping and brightness shifts. For more severe domain gaps (e.g., inserting a half-body reference into a full-body mask), we experiment with reference cropping. While easing the gap, random cropping also introduces some hallucinations and detail loss; we therefore disable reference cropping in the final training as a trade-off.

\noindent\textbf{Two-phase Training Scheme.} As mentioned above, the pre-training dataset only uses references extracted from their source videos, limiting generalization to external references with deformations, brightness discrepancies or domain gaps. We address this with a two-phase training:

(i) \emph{Pre-training (self-attention only):} We first train the base model on our core HumanVID-derived dataset (Sec. 3.2) with only the self-attention layers learnable, preserving the foundational generative capabilities of the base model.

(ii) \emph{Quality tuning (full fine-tuning):} We assemble a smaller, high-quality dataset of diverse reference-video pairs by (a) extracting paired images from AnyInsertion \cite{insert} and Subject200K \cite{subject200k} and converting one image of each pair to video via Wan-I2V \cite{wan}, and (b) adding $\sim$400 handheld object pairs from AnchorCrafter-400 \cite{anchorcrafter}. All the model layers are unfrozen for full fine-tuning, enabling better convergence and adaptation to cross-domain reference-video pairs.

\noindent\textbf{Length Extension and First Frame Reference.} Our design (Sec. 3.2, dummy reference latent) provides an interface for handling long videos and leveraging first-frame priors.

(i) \emph{Long video processing:} For videos longer than our training length, we first split the video sequence into overlapping segments: the last frame of the previous segment becomes the first frame and the dummy reference for the next. In subject swapping, the majority of scene content remains constant between segments. This inherent context enables stable temporal extrapolation without the disruptive scene jumps common in general image-to-video tasks.

(ii) \emph{First frame reference:} Since the dummy reference concentrates model's attention on the first frame, any edit applied there propagates coherently. At inference, users may optionally supply a first‑frame swap produced by an image subject swapping model like \cite{anydoor}, enhancing visual fidelity. Notably, DreamSwapV is inherently compatible with single frame swapping, so leveraging specialized image-based models is an alternative option for stability.

\noindent\textbf{Tunnel Video Inpainting and Reweighting Loss.} When the mask is extremely small (e.g., accessory), the model may overlook fine details. Inspired by Tunnel Try-on \cite{tunnel} and AnchorCrafter \cite{anchorcrafter}, we tackle this issue in parallel during both inference and training:

(i) \emph{Inference—tunnel inpainting:} At inference time, when the mask area ratio falls below the predefined threshold (0.05), we extract a tight crop (\emph{tunnel}) around the mask, perform subject swapping in this sub-region, and blend this tunnel region back into the full frame, which can focus on pixels where it matters, improving detail fidelity on tiny objects.

(ii) \emph{Training—reweighting loss:} Following AnchorCrafter, we enhance the model's attention to small subject regions by introducing a subject-region reweighting loss:
\begin{equation}
\begin{aligned}
L_{\text{rw}}
  &= 
     \frac{\mathbf{E}}{\mathbf{E}^{s}}\,
     \mathbf{M}^{s} \odot L_{\text{pt}}
     \\[4pt]  
L_{\text{final}}
  &= \bigl(1 - \mathbf{M}^{s}\bigr)\odot L_{\text{pt}}
     \;+\;
     \lambda L_{\text{rw}}
\end{aligned}
\end{equation}

where $L_{\text{pt}}, L_{\text{final}}, L_{\text{rw}}$ are pre-trained loss, overall training objective, and re-weighting loss that amplifies the learning signal inside the subject mask region; $\mathbf{E}$ and $\mathbf{E}^s$ are the areas of the whole frame and the subject, respectively, and $\mathbf{M}^{s}\in \{0,1\}^{h\times w}$ indicates the subject binary mask.

\begin{figure*}[]
\centering
\includegraphics[width=0.95\textwidth]{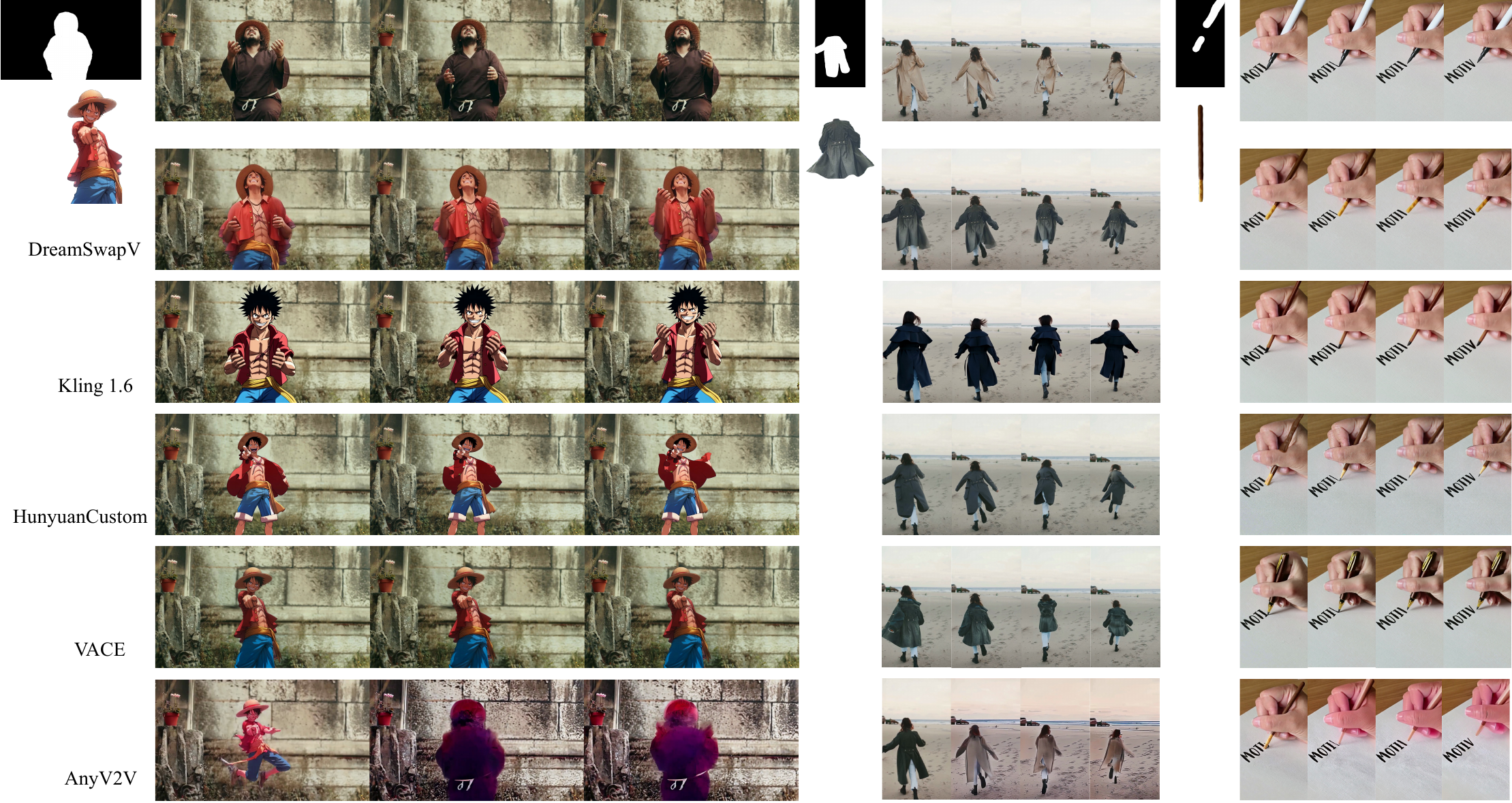} 
\caption{Qualitative comparisons between our DreamSwapV and four baselines across diverse subjects and video aspect ratios. Please zoom in for details, and see the supplementary video for dynamic results.}
\label{quanti}
\vspace{-1.5em}
\end{figure*}
\section{Experiments}

\subsection{Baselines and Benchmark}

\noindent\textbf{Basic Information.} Our DreamSwapV is trained based on Wan-I2V-14B foundation model, supporting up to 720p resolution. Our training follows a two-phase scheme: pre-training for 15000 iterations and quality-tuning for 10000 iterations, both on 32 NVIDIA H100 GPUs. Each 100 iterations cost approximately 75 minutes, resulting in a total training duration of $\sim$12.5 days to obtain the final model.

\noindent\textbf{Benchmark.} While VBench \cite{vbench} and VBench++ \cite{vbench++} have provided valuable benchmarks for text-to-video and image-to-video tasks, similar evaluations for video subject swapping remains absent. To address this gap, we introduce DreamSwapV-Benchmark, the first benchmark dedicated to evaluating video subject swapping. Following the scale of benchmarks used in VACE and HunyuanCustom, we collect 100 videos from the online website \cite{pexels}, and annotate 167 distinct subject instances from them. Original videos cover four aspect ratios to test cross‑resolution generalization. For each subject, we segment and track its precise mask sequence in the source video, and match it with an appropriate reference image for swapping. The subjects and their reference images for swapping are carefully selected to ensure a wide and balanced distribution across categories and complexities.

We adopt five indicators inherited from VBench (see the top 5 metrics in Table \ref{table}) and design two other automatic metrics specifically for the subject swapping task: reference appearance and background preservation. We also introduce a user study to evaluate human preferences for three aspect: reference detail, subject interaction and visual fidelity. The full benchmark distribution, metric calculation rules and user study details are provided in Appendix 3.2.

\noindent\textbf{Baselines.} We compare our DreamSwapV against 3 open-source methods (AnyV2V, VACE, HunyuanCustom) and 1 commercial model (Kling 1.6 Multimodal). We select them for their robust stability, high accessiblity, and close relevance to our subject swapping objective. We exclude text-instruction based methods (VideoGrain, VideoSwap, VideoPainter) to avoid unfair comparisons under distinct task settings. The specific implementations and adaptations for the four baselines are detailed in Appendix 3.3.

\subsection{Evaluation}

\noindent\textbf{Quantitative Comparison.} Table \ref{table} reports the quantitative results on DreamSwapV‑Benchmark. Across the five VBench indicators, our DreamSwapV achieves the best average score, marginally surpassing Kling 1.6. The gap widens when Kling 1.6 suffers from its \emph{regeneration-like} framework, altering large portions of the background—or even redrawing entire frames—compromising its practical usability. VACE and HunyuanCustom underperform mainly in reference detail and dynamic degree, occasionally inserting incorrect or static subjects due to their unified task focus, where subject swapping is only a part component. AnyV2V's unstable intermediate feature manipulations often lead to complete video collapse, severely impairing visual fidelity. Overall, DreamSwapV delivers the most stable and high-quality results, corroborated by the user study.

\noindent\textbf{Qualitative Results.} Figure \ref{quanti} provides a visual comparison highlighting DreamSwapV's advantages across 3 different kinds of subjects (see Appendix 4.2 for more results). In the \emph{Human$\Rightarrow$Luffy} case, DreamSwapV successfully swaps the person with Luffy, preserving his original pose and reference details, while Kling struggles with reference consistency, HunyuanCustom and VACE injects an almost static subject, and AnyV2V collapses completely. In the \emph{Coat$\Rightarrow$Coat} and \emph{Pen$\Rightarrow$Pocky} case, all baselines fail to maintain the reference appearance, whereas DreamSwapV retains fine details and realistic motion, yielding the highest visual quality.

\noindent\textbf{More applications.} It is worth noting that DreamSwapV is not limited to video subject swapping, but can extend to several related applications like \emph{image swapping, video inpainting, video addition, video try-on} and so on. We further discuss these extensions, along with limitations in Appendix 5.

\begin{figure}[t]
\centering
\includegraphics[width=0.48\textwidth]{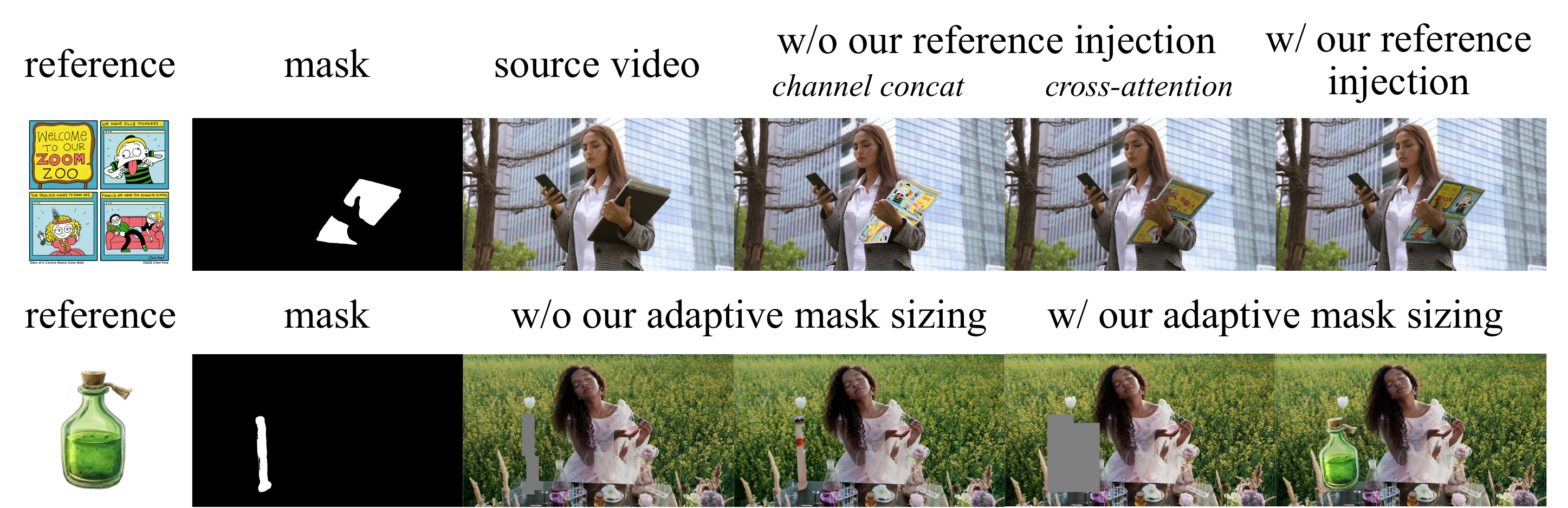} 
\caption{Representative ablation results on our reference injection and adaptive grid sizing, zoom in for details.}
\label{abla}
\vspace{-1.5em}
\end{figure}
\subsection{Ablation Study}

We ablate on our reference injection of condition fusion module and adaptive grid sizing with visual examples in Figure \ref{abla}. Our reference injection achieves finest details over alternatives like \emph{channel concatenation} and \emph{cross-attention}. Our adaptive grid sizing can better handle subjects with different scales and improve visual fidelity. More quantitative ablation results can be found in Appendix 4.1.

\section{Conclusion}

In this paper, we present DreamSwapV, a mask-guided, subject-agnostic framework, dedicated to end-to-end generic subject swapping for video customization. We design a condition fusion module and an adaptive mask strategy specifically for the task, efficiently integrating various conditions for more fine-grained control and improving interactions between diverse subjects and their surroundings. Our elaborate dataset construction and two-phase training scheme further amplify our DreamSwapV's capabilities, enabling it to outperform all existing methods on VBench indicators and our first introduced DreamSwapV-Benchmark.

\bibliography{aaai2026}

\end{document}